\documentclass[a4paper, 10pt, twocolumn, twoside]{article}
\usepackage{ISARC}
\usepackage{adjustbox}
\usepackage{lscape}
\usepackage{hologo}
\usepackage[utf8]{inputenc}
\usepackage{float}
\usepackage[linesnumbered,ruled,vlined]{algorithm2e}
\usepackage{booktabs}
\usepackage{algpseudocode}
\usepackage{tabularx}

\usepackage[english]{babel}
\usepackage[autostyle, english = american]{csquotes}
\MakeOuterQuote{"}

\begin{document}

 % Do not change the following line
\linespread{0.5}

\title{Bayesian BIM-Guided Construction Robot Navigation with NLP Safety Prompts in Dynamic Environments}

\author{Mani Amani$^{1 , 2}$ and Reza Akhavian$^1$}

\affiliation{
$^1$Department of Civil, Construction, and Environmental Engineering, San Diego State University, United States\\
$^2$Department of Electrical and Computer Engineering, University of California San Diego, United States\\
}

\email{
\href{mamani@ucsd.edu}{mamani@sdsu.edu}, 
\href{rakhavian@sdsu.edu}{rakhavian@sdsu.edu}
}

% Do not change the following three lines
\maketitle 
\thispagestyle{fancy} 
\pagestyle{fancy}

\begin{abstract}
Construction robotics increasingly relies on natural language processing for task execution, creating a need for robust methods to interpret commands in complex, dynamic environments. While existing research primarily focuses on what tasks robots should perform, less attention has been paid to how these tasks should be executed safely and efficiently. This paper presents a novel probabilistic framework that uses sentiment analysis from natural language commands to dynamically adjust robot navigation policies in construction environments. The framework leverages Building Information Modeling (BIM) data and natural language prompts to create adaptive navigation strategies that account for varying levels of environmental risk and uncertainty. We introduce an object-aware path planning approach that combines exponential potential fields with a grid-based representation of the environment, where the potential fields are dynamically adjusted based on the semantic analysis of user prompts. The framework employs Bayesian inference to consolidate multiple information sources: the static data from BIM, the semantic content of natural language commands, and the implied safety constraints from user prompts. We demonstrate our approach through experiments comparing three scenarios: baseline shortest-path planning, safety-oriented navigation, and risk-aware routing. Results show that our method successfully adapts path planning based on natural language sentiment, achieving a 50\% improvement in minimum distance to obstacles when safety is prioritized, while maintaining reasonable path lengths. Scenarios with contrasting prompts, such as "dangerous" and "safe," demonstrate the framework’s ability to modify paths based on. This approach provides a flexible foundation for integrating human knowledge and safety considerations into construction robot navigation.
\end{abstract}

\begin{keywords}
Construction Robotics, Natural Language Processing, Bayesian Inference, Robotic Path Planning, Building Information Modeling (BIM), Sentiment Analysis, Exponential Potential Fields, Dynamic Environments, Robot Safety
\end{keywords}

\begin{figure}
    \centering
    \includegraphics[width=1\linewidth]{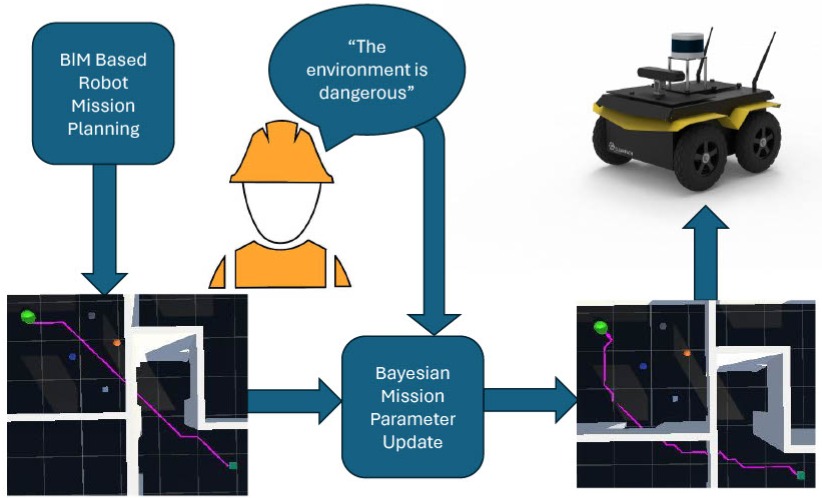}
    \caption{Semantic Reasoning using BIM and Natural Prompts to Update Robotic Tasks}
    \label{fig:main}
\end{figure}
\section{Introduction}
\label{sec:Introduction}

Construction jobsites present unique challenges for autonomous robot navigation due to their inherently dynamic and cluttered nature. Unlike controlled manufacturing environments, construction sites undergo constant changes as work progresses, with temporary structures, moving equipment, materials, and workers creating a complex and evolving workspace. While BIM provides valuable spatial and semantic information for robot planning, its static nature may not fully capture the dynamic reality of construction environments \cite{[B]}. The discrepancy between BIM data and actual site conditions, combined with the presence of temporary obstacles and moving objects not represented in the BIM, creates significant challenges for safe and efficient robot navigation \cite{[C]}. These challenges are particularly acute in renovation projects where existing conditions may differ from available BIM data \cite{[D]}. To address this gap, there is a growing need for adaptive path-planning algorithms that can incorporate both the static spatial context from BIM and dynamic real-time information. Furthermore, integrating probabilistic methods, such as Bayesian inference, offers a pathway to manage uncertainty and variability through unsafe event prediction \cite{SIMAANBayesian}.

\par

Robot planning with natural language processing (NLP) has received significant attention in recent years \cite{UCSDLLM}.
The construction industry presents a unique opportunity to use BIM, digital twins, and the rich textual and contextual information they contain to leverage the advancements in robot planning with NLP \cite{LLMBIMreview}. Some of the recent promising studies look for predefined mission keywords and commands to execute the intended task \cite{VineetLLM}. Other work has shown that using LLMs to identify object functions can result in misinterpretations and mission failure \cite{FailLLM}. The embedded information in BIM includes a rich spatiotemporal and textual database regarding the mission environment, making them attractive candidates to incorporate into mission planning \cite{BIMrobotplanning}.  Previous works have leveraged schedules, pose estimation, and path planning information from BIM for robotic applications \cite{BIMestimationrobot}. However, the fidelity of the information representation is crucial to ensure robust real-world implementations, as there might be inaccuracies in the model data. Such errors can cause robot malfunction and poor mission planning for autonomous agents. Therefore, real-time simulation and representation of a 3D map and accurate BIM localization is an active research topic in the construction research field \cite{REaltimeBIMLocal}. 
\par
Moreover, the current research on language-commanded construction robotics is limited to predefined objects and tasks. Both learning-based methods and hard-coded approaches can cause significant compatibility problems in the long run due to different nomenclatures from project A to project B and from one BIM Execution Plan to another. Furthermore, these approaches do not consider the prompter's authority, knowledge, and reliability. In many cases, the prompter is assumed to be completely reliable and have full authority over the robot's decision-making. However, in practical applications, it is important to distinguish the rank and the confidence level of the given prompt. Given that most neural networks are trained on large datasets, unseen commands can cause inaccuracies and misunderstandings that can harm robot operations \cite{LLMReviewPAPERROBOT}. It is important to identify the confidence and relevance level of prompts to the robot to ensure safe mission planning. To advance upon these shortcomings, we propose a probabilistic framework that uses sentiment analysis from commands to adjust the behavior of the mission given an end goal. Figure \ref{fig:main} depicts the overall idea of the paper. 
\par
Bayesian inference has been one of the cornerstones of probability theory. From economics to healthcare, many fields have used Bayesian inference to make predictive decisions from prior and observed data \cite{BayesinHealthcare}, and it has been used extensively in construction planning and decision-making \cite{SIMAANBayesian, MiroslawBayesian}. We propose to use Bayesian updates to consolidate current danger evaluations of the environment with the sentiment received from the prompt for more accurate and dynamic robotic pathfinding which is imperative in cluttered environments such as construction jobsites. 

\section{Problem Setup}
We introduce a framework to extract arbitrary sentiments from user prompts within the parameters of BIM families. While this sentiment analysis method is generalizable to any problem that could use language consolidation, we demonstrate its efficacy in solving robotic pathfinding. The initial iteration of the NLP method uses the family names within the BIM to scale repellent coefficients associated with this object. This coefficient determines the "danger level", and can represent dynamism, localization error, cost, and any other heuristic the user chooses to use to define which family of objects should be repulsive for the robot to avoid getting close to in the context of path planning. The next step is to incorporate this information as a natural language command. The received command is then integrated by a transformation into numerical values of the sentiment of the user about the danger of the environment. For example, if the prompt mentions that the environment is dangerous (i.e., be careful when you go into room A) or conversely, the environment is safe (i.e., room A should be empty, go quickly), we can map these into a danger coefficient ranging from 0-1. These coefficients will then be used to scale any chosen heuristic that would alter this behavior. We expand on this more in the following sections.

\section{Mathematical Framework}
\subsection{Exponential Potential Field (EPF)}
The distance to the nearest obstacle is a widely used metric for object avoidance in robot navigation \cite{Bubblecover}. Common approaches such as signed distance fields and Euclidean distance fields are processes to represent space and occupancy using the closest distance of every obstacle relative to the current coordinates \cite{SDF}. The distance field can be interpreted as a "potential force" emitted from objects within the traversable areas. This approach was pioneered with the development of artificial potential fields (APF) \cite{APF}. APFs are a series of attractive and repulsive potentials that enable path planning for autonomous agents by traversing an environment using repulsive forces from obstacles and an attractive force from the goal. Usually, the repulsive function in these settings is formulated as an exponential function that exponentially increases in value as the agent gets closer to an obstacle, resulting in a stronger repulsive force. In this work, we employ a similar exponential function resulting in an "Exponential potential field" or EPF. The reason we use an EPF instead of an APF is that we do not need an attractive potential since we use a graph traversal approach which does not need an attractive potential for navigation.
 \par
  The motivation behind using EPFs is to account for slight variations between the reconciling of BIM and the real world. The exponential function creates a stronger force for areas closer to obstacles. Furthermore, we formulate a cumulative metric in the form of the summation of all potential values on each grid point using Equation \ref{eq:grid_point_sum} as opposed to the minimum distance done in traditional distance field approaches. This cumulation is more akin to APF for navigation. In complex and dense environments, the cumulative metric tends to result in higher potential values, which naturally encourages paths to avoid these areas whenever possible. We then discretize the map into a grid form to be able to use graph traversal methods for path planning. The calculation of the EPF is shown in Equation \ref{eq:repulsive_force}:
\begin{equation}
f_{\text{rep}}(x, y, \mathcal{M}) =
\begin{cases} 
k_{\text{rep}} e^{-D(x, y, \mathcal{M})}, & \textstyle D(x, y, \mathcal{M}) < D_{max}\\
0, & \textstyle \text{otherwise.}
\end{cases}
\label{eq:repulsive_force}
\end{equation}
where \(\mathcal{M}\) is the set of points \((x, y)\) that are the point of collision of the 3D object with the 2D plane and \(k_{\text{rep}}\) is a scalar that scales the potential value. Equation \ref{eq:repulsive_force} will equal \(k_{\text{rep}} e^{-D(x, y, \mathcal{M})}\) when the distance (calculated using Equation \ref{eq:distance}) is lower than \(D_{max}\), which is a threshold to reduce extremely small potential values to ensure numerical stability, and 0 otherwise.
\begin{equation}
D(x, y, \mathcal{M}) = \min_{(x', y') \in \mathcal{M}} \sqrt{(x - x')^2 + (y - y')^2}
\label{eq:distance}
\end{equation}

Each grid point will have a cumulative potential value represented by Equation \ref{eq:grid_point_sum}:
\begin{equation}
    G(x_i, y_j) = \sum_{k=1}^{O} f_{\text{rep}}(x_i, y_j, \mathcal{M}_k), \quad \forall (x_i, y_j) \in  \mathcal{S}
\label{eq:grid_point_sum}
\end{equation}

\(G(x_i,y_i)\) is then used as a heuristic for our path-finding regimen. 
\subsection{Multi Heuristic A*}
A* search algorithm is a graph traversal and pathfinding algorithm that is used in robotics and computer science \cite{russell2016artificial}. The algorithm uses a set of weighted graphs to find the most optimal path given a heuristic. At its core, A* relies on the calculation of cost at each node given by Equation \ref{A*}

\begin{equation}
    f(n) = g(n) + h(n)
    \label{A*}
\end{equation}
where \(f(n)\) is the total cost of each node, \(g(n)\) is the cost of the start node to the current node, and \(h(n)\) is the heuristic of the current node.
\par
Given an accurate and admissible heuristic, the algorithm is guaranteed to find the optimal path. Admissibility in the context of heuristics refers to a heuristic never overestimating the cost; this is usually determined by the triangle inequality. In other words, the admissibility of the heuristic guarantees A* will never overestimate the cost of each node and returns the optimal path \cite{MHA*}.  However, for complex tasks, it is notoriously difficult and at times impossible to formulate a single admissible heuristic. \cite{MHA*}. This is the main motivation behind the creation of the multi-heuristic A* (MHA*). MHA* algorithmically handles multiple heuristics around one admissible anchor heuristic \cite{MHA*}. 
\par
In this problem, we intend to use both distance and EPF as complementing heuristics for each node. However, the EPF can be potentially inadmissible. Therefore, we use one admissible heuristic in the form of Euclidean distance and one potentially inadmissible heuristic in terms of the EPF. While using inadmissible heuristics sacrifices the optimality guarantee of A*, MHA* can still ensure optimality by employing an admissible anchor heuristic, even when other heuristics are potentially inadmissible. In this scenario with EPF and Euclidean distance, MHA* will use both heuristics to calculate the most optimal path.
\label{sec:Preparation}
\subsection{Bayesian Inference of NLP Sentiment Analysis}

We propose to map the safety sentiment of the BIM families and the NLP prompts to a probabilistic value using LLM reasoning capabilities. The probabilistic value ranges from 0 to 1, and the higher number suggests a higher possibility of dynamism, value, and localization errors due to object geometry or location relative to the sensors. A previous work by the authors has shown the validity of using GPT risk sentiment analysis concerning BIM families \cite{UsAPF}

Initially, an EPF is generated using the BIM families present in the model where the robot is planned to be used, using GPT-produced coefficients to scale EPF values. This value will serve as the prior for each of the objects' EPF scaling factors. The user's prompt will serve as evidence in the Bayesian inference framework. This gives us a robust mathematical framework to be able to update our scalars. Bayesian inference is given by Equation \ref{Bayes equation}:

\begin{equation}
    P( H | E) = \frac{P(E | H) P(H)}{P(E)}
    \label{Bayes equation}
\end{equation}
where, \(E\) is the evidence or observed state, and \(H\) is the hypothesis or our prior.

In this case, we treat our prior hypothesis \(H\) as the initial output of the GPT using BIM families. The evidence \(E\) will be the output of the GPT given the prompt and the prior information given previous outputs. Since our distribution is discrete given that the commands are an integer number of prompts, we can expand \(P(E)\) in the following form.
\begin{equation}
    P( H | E) = \frac{P(E | H) P(H)}{P(E | H) P(H) + P(E | \neg H) P(\neg H)}
    \label{eq5}
\end{equation}
The term \(P(E | \neg H) P(\neg H)\) in the denominator is problem-dependent. For simplicity, one can interpret the denominator as a scaling constant that can be tuned to the problem constraints.
\par
The formulation in Equation \ref{eq5} allows us to use infinitely many prompts to update our path-finding landscape.

\begin{equation}
    P(H \mid E_1 \ldots, E_n) = 
    \frac{P(H) \prod_{i=1}^n P(E_i \mid H, E_1, \ldots, E_{i-1})}{P(E_1, E_2, \ldots, E_n)}.
    \label{intractible}
\end{equation}

Equation \ref{intractible} can generally be intractable. This is due to the conditional probability term in the numerator that is notoriously difficult to evaluate. In practice, to simplify this computation, we assume that the likelihoods are conditionally independent. This will simplify the problem even further:

\begin{equation}
    P(H \mid E_1 \ldots, E_n) = 
    \frac{P(H) \prod_{i=1}^n P(E_i \mid H)}{P(E_1, E_2, \ldots, E_n)}.
\end{equation}

Where:
\begin{equation}
    P(E_1, E_2, \ldots, E_n) = \sum_{H'} P(H') \prod_{i=1}^n P(E_i \mid H').
\end{equation} 
Yielding the final form of:

\begin{equation}
    P(H \mid E_1 \ldots, E_n) = 
    \frac{P(H) \prod_{i=1}^n P(E_i \mid H)}{\sum_{H'} P(H') \prod_{i=1}^n P(E_i \mid H')}.
\end{equation}

\begin{figure}[htbp]
    \centering
    \begin{subfigure}{1\columnwidth}
        \centering
        \includegraphics[width=\linewidth]{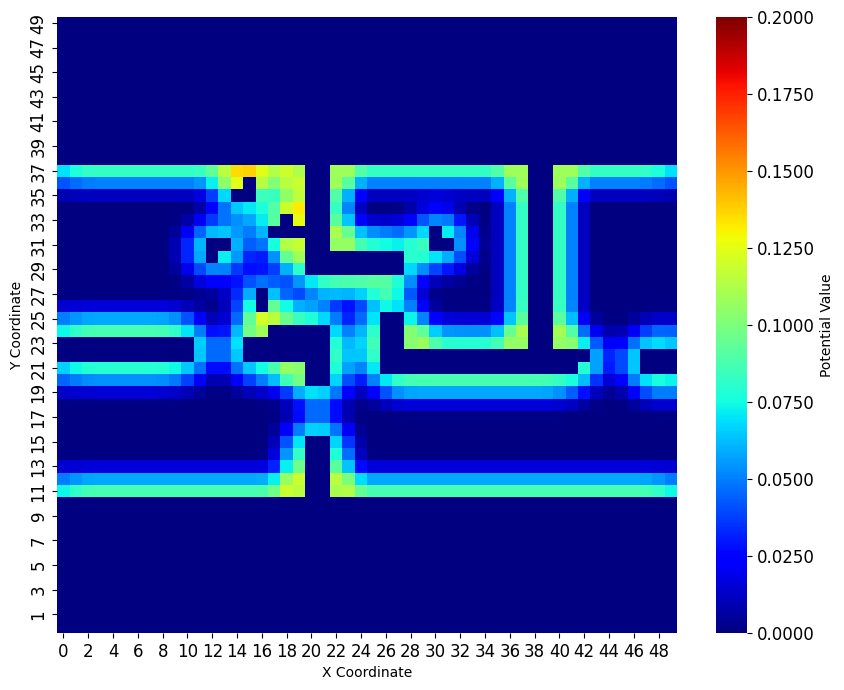}
        \caption{Without using any prompts}
        \label{fig:prebayes_dangerous}
    \end{subfigure}
    \hfill
    \begin{subfigure}{1\columnwidth}
        \centering
        \includegraphics[width=\linewidth]{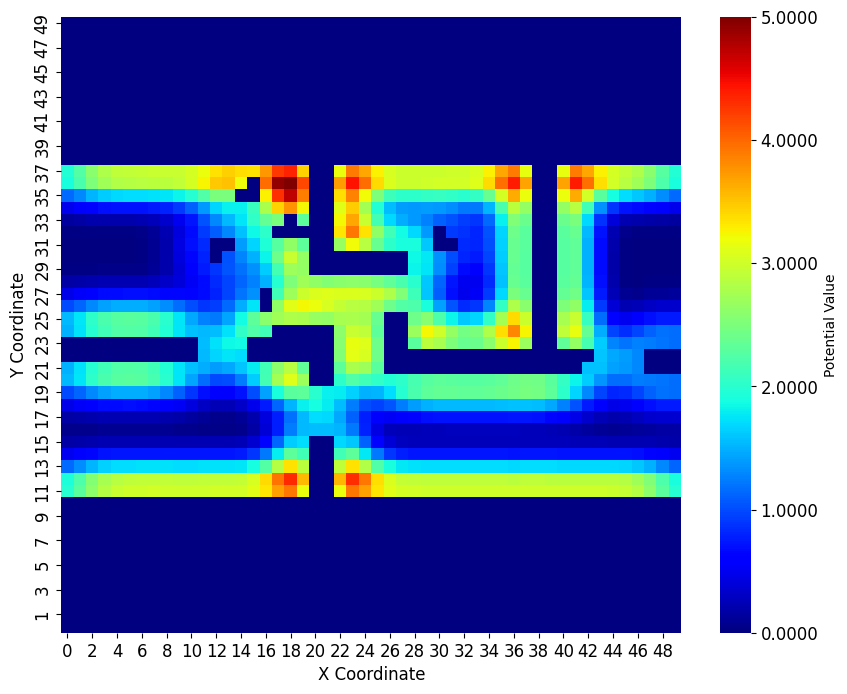}
        \caption{With a prompt implying a dangerous environment}
        \label{fig:postbayes_dangerous}
    \end{subfigure}
    \hfill
    \begin{subfigure}{1\columnwidth}
        \centering
        \includegraphics[width=\linewidth]{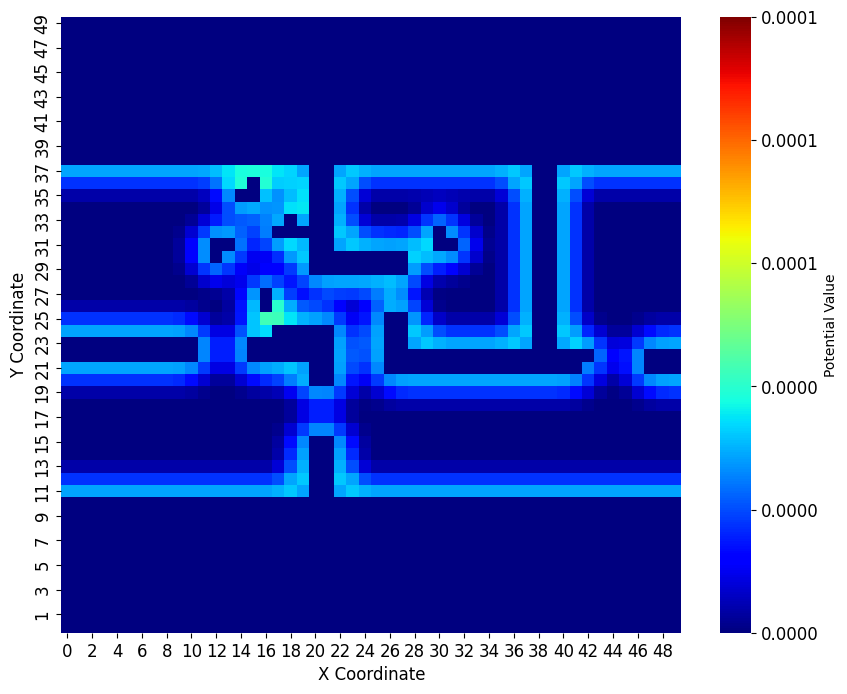}
        \caption{With a prompt implying a safe environment}
        \label{fig:postbayes_safe}
    \end{subfigure}
    \caption{Comparison of results under different conditions: (a) without using any prompts, (b) using a prompt that implies a dangerous environment, and (c) using a prompt that implies a safe environment.}
    \label{fig:comparison_prompts}
\end{figure}

This formulation will allow us to chain multiple language commands with varying confidence ratios with our planning framework. Once the ultimate coefficient \(P(H \mid E_1 \ldots, E_n)\) is calculated, we can use it directly as the \(k_{rep}\) or \(D_{max}\) term. in Equation \ref{eq:repulsive_force}. Since the prompts change the environment's potential value, each node's potential heuristic can be also changed. Figure 
\ref{fig:comparison_prompts} shows an example of the effects on the potential field given different prompts. The potential value of each grid is altered by the framework's analysis of prompt safety. The heatmap in Figure \ref{fig:comparison_prompts} denotes safe to dangerous areas using a blue-to-red color spectrum, and the values designated to the colors are the value of the EPF on each grid point, starting from 0 for the safest situation to 5 for the most dangerous. Figure \ref{fig:prebayes_dangerous} represents the initial evaluation of the scene that follows neutral scaling factors generated from the GPT and BIM families. However, in 
Figures \ref{fig:postbayes_dangerous} and \ref{fig:prebayes_dangerous} we see the effects of the prompt on the values of the potential field. When the framework receives a prompt with dangerous sentiment, the potential field increases in strength, which affects the path by choosing a longer but safer path. Conversely, when the framework is represented with a safer prompt, the potential field decreases in strength as seen in Figure \ref{fig:postbayes_safe}. Furthermore, when the potential field is zero across the map, the pathfinding problem reduces to the naive A* algorithm. It is important to note that this probabilistic update is not limited to this specific problem formulation. Any function or heuristic can have its parameters altered given these updates, resulting in different policy realizations. 
\section{Methodology}
\begin{figure}[h]
    \centering
    \includegraphics[width=0.75\linewidth]{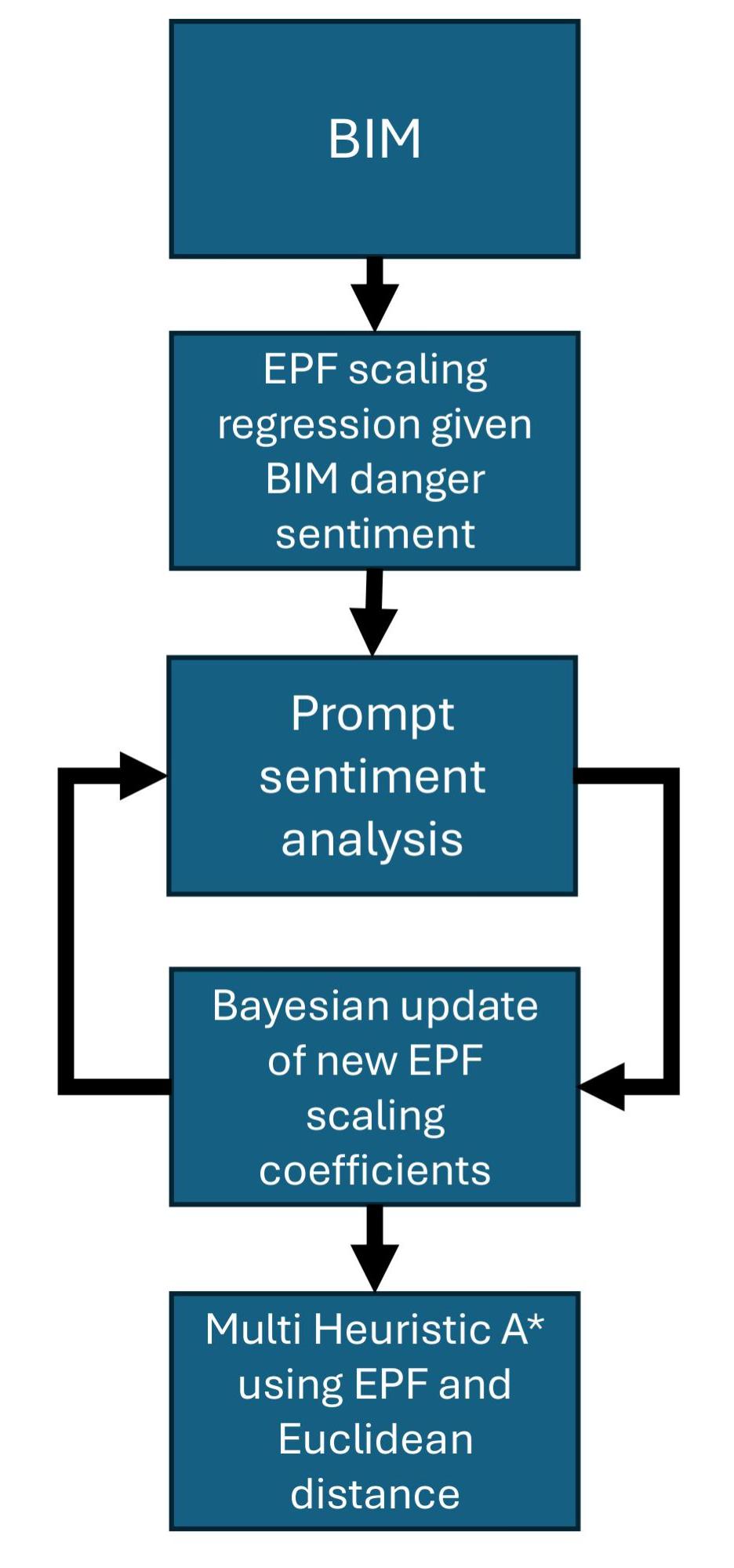}
    \caption{NLP processing framework}
    \label{fig:framework}
\end{figure}

The developed framework loads a BIM in a robot simulation platform such as Unity or Gazebo in the form of an FBX file. Once the BIM has loaded, the names of the families are parsed and prepared in a JSON file for further processing. The JSON file is then prompted into any LLM platform with an associated prompt. Each object is then associated with a specific scalar given the LLM's reasoning regarding the regression value from the sentiment of the family's name with the context of the rest of the BIM. These coefficients are available to be used in the context of scaling the EPF. Given a prompt or a chain of prompts, the framework can start updating the coefficient. The prompt is appended with the current state \(P(H)\) to yield current conditioned evidence \(P(E|H)\). The new information is parsed from the response of the GPT. We then calculate the new coefficients using Equation \ref{eq5}.
Once the coefficients are retrieved, the EPF is recalculated and ready to be used with MHA* for robotic planning. Figure \ref{fig:framework} illustrates this process.
\par
\section{Experiments \& Results}
\begin{figure}[H]
    \centering
    \includegraphics[width=1\linewidth]{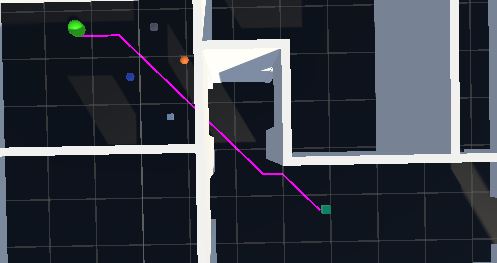}
    \caption{Baseline Path Calculated by A*}
    \label{fig:A* baseline}
\end{figure}

\begin{figure}[H]
    \centering
    \includegraphics[width=1\linewidth]{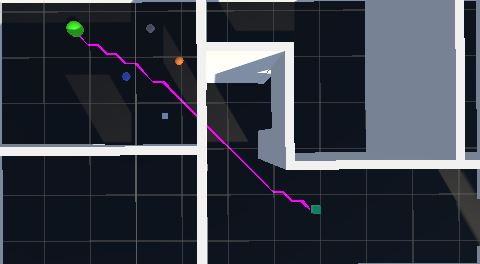}
    \caption{Calculated path given safe prompt}
    \label{fig:safe}
\end{figure}

\begin{figure}[H]
    \centering
    \includegraphics[width=1\linewidth]{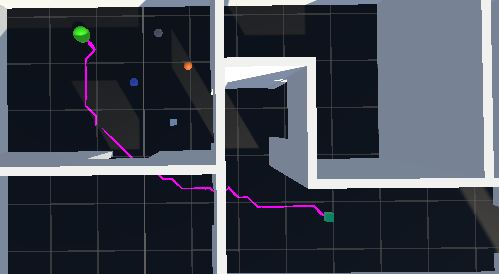}
    \caption{Calculated path given dangerous prompt}
    \label{fig:danger}
\end{figure}

We consider the shortest path possible and the average Euclidean distance from any given environment as our baseline to evaluate the efficacy of the method. This value is returned by the classical A* algorithm which guarantees the shortest path. The path can be seen in Figure \ref{fig:A* baseline}.
We then experiment with the same starting point using two different prompts. While our theoretical framework allows for the consolidation of infinitely many prompts, we examine the single prompt setting and leave the analysis of the multi-prompt scenario for future works. We choose two extremes to demonstrate the prompt's effect on the path. One extreme would be "The environment is incredibly safe" and "The environment is incredibly dangerous" for the other extreme. Figures \ref{fig:safe} and \ref{fig:danger} show each condition respectively. There is a clear qualitative and quantitative difference in the length and behavior of the path given simply a different prompt regarding the context of the prompt.
Table \ref{tab:pathastats} shows the statistics and improvements in speed and minimum distance to obstacles (MDO). 
\par
The baseline algorithm implementation will always result in the shortest path possible due to the guarantees regarding optimizing the single heuristic. In this case, the heuristic is Euclidean distance which will result in the shortest path. This approach will not take obstacle distance into account. The multi-heuristic approach, however, enables the ability to do so, as reflected in Table \ref{tab:pathastats}. We can see that the safe prompt improves the obstacle avoidance metric at a small cost in path length. This is due to a small potential field being generated as a consequence of a safe prompt. The effects of the prompt on the coefficients that affect the EPF are shown in Tables \ref{tab:stats2} and \ref{tab:statsdangeours}. Furthermore, we can see a large improvement in the object avoidance metric when the framework is presented with a prompt implying a dangerous environment, even though this comes at the cost of taking a longer path. This framework enables users to be able to both optimize between distance and object avoidance while being able to inform the robot of the world state.
\begin{table}
    \centering
        \caption{Bayesian path metrics for all scenarios}
    \begin{tabular}{@{}lcc@{}}
    \toprule
    Strategy   & Path Length(m) & MDO(m) \\ \midrule
    Baseline               &\textbf{ 3.286}          & 0.40 \\ \midrule
    Safe &    3.288  &  0.43\\\midrule
    Dangerous & 3.922   & \textbf{0.60} \\
    \bottomrule
    \end{tabular}
    \label{tab:pathastats}
\end{table}

\begin{table}[h]
    \centering
        \caption{Bayesian update safe scenario}
    \begin{tabular}{@{}lccc@{}}
    
    \toprule
    BIM family   & Prior Coef. & New Coef. & Updated Coef.\\ \midrule
    Wall         & 0.2 & 0.02 & 0.03\\ \midrule
    Grinder &    0.8&  0.08 & 0.78\\\midrule
    Chainsaw &   0.95  &  0.09& 0.90\\\midrule
    Robot &    0.9  &  0.09& 0.91\\\midrule
    Chair & 0.6 & 0.06 & 0.21 \\
    \bottomrule
    \end{tabular}

    \label{tab:stats2}
\end{table}

\begin{table}[h]
    \centering
        \caption{Bayesian Update dangerous Scenario}
    \begin{tabular}{@{}lccc@{}}
    \toprule
    BIM family   & Prior Coef. & New Coef. & Updated Coef.\\ \midrule
    Wall         & 0.3 & 0.8 & 0.01 \\ \midrule
    Grinder &    0.7 &  0.9 & 0.59\\\midrule
    Chainsaw &    0.9  &  1& 0.59\\\midrule
    Robot &    0.8  &  1 & 0.51\\\midrule
    Chair & 0.1  & 0.6 & 0.43 \\
    \bottomrule
    \end{tabular}

    \label{tab:statsdangeours}
\end{table}

\section{Discussion \& Limitations}
The described methodology presents a computationally efficient and transparent opportunity for consolidating and analyzing sentiment from NLP commands. However, its integration with different planning strategies remains to be explored. Currently, the integration is with a heuristic-based approach, meaning that the algorithm finds an optimal policy given the heuristic of choice. This may or may not be effective given different conditions and prompts. For example, if the space only has one possible route from the starting point to the goal point, no matter how strong the prompt and by extension the EPF might be, the only viable solution would be that route. The optimal policy can remain unchanged if the prompt provides slight differences in heuristic parameters. Other approaches must be considered to integrate Bayesian consolidation into a more granular and flexible planning strategy.
\par
Furthermore, as with the state-of-the-art LLMs, there is always a possibility of hallucinations, which yields inaccuracies such as incorrect likelihood coefficients. To quantify the reliability of GPT 3.5-turbo for this task, we experimented with two prompts and a baseline evaluation with no prompts over 100 iterations to measure the statistics of how consistent it is when prompted to
perceive semantic information. Table \ref{tab:GPTstats} represents the statistics regarding GPT 3.5-turbo output stability. In this table, the values represent the danger levels associated with each BIM family when no prompts are given, with a prompt implying a dangerous environment, and with a prompt implying a safe environment. It can be seen that the GPT performs reasonably well in assigning new danger values to most BIM objects with respect to the sentiment of the prompt. This comparison over several iterations indicates that the likelihood of false or inaccurate coefficients is relatively low in simple prompts and NLP commands such as those needed for the proposed framework to generate the expected results. While GPT 3.5-turbo is a fast and cost-effective choice, it lacks the advanced reasoning capabilities of more modern counterparts such as GPT 4 \cite{GPTbad}. To ensure better parameter scaling and lower variance, using more advanced GPT models such as GPT 4o or GPT o1 is recommended.
\begin{table}[h!]
    \centering
        \caption{Danger Levels from 100 iterations to assess the reliability of GPT 3.5-turbo in understanding prompt semantics}
    \begin{tabular}{@{}p{1.7cm}p{1.7cm}p{1.7cm}p{1.7cm}@{}}
    \toprule
    BIM family & Original Value & Safe Prompt & Dangerous Prompt \\ \midrule
    Wall & 0.27 \(\pm\) 0.18 &             0.18 \(\pm\) 0.17 &          0.43 \(\pm\) 0.31\\ \midrule
    Grinder     & 0.71 \(\pm\) 0.15 &             0.60 \(\pm\) 0.33 &           0.74 \(\pm\) 0.19 \\ \midrule
    Robot       & 0.60 \(\pm\) 0.11 &         0.50 \(\pm\) 0.28 &       0.67 \(\pm\) 0.18 \\ \midrule
    Chainsaw    & 0.74 \(\pm\) 0.17 &          0.61 \(\pm\) 0.34        & 0.75 \(\pm\) 0.19\\ \midrule
    Chair       & 0.32 \(\pm\) 0.14 &           0.21 \(\pm\) 0.15 &                0.44 \(\pm\) 0.27 \\ \bottomrule
    \end{tabular}
    \label{tab:GPTstats}
\end{table}

\section{Conclusion and Future Works}
This paper proposes a new approach to consolidate and use natural language prompts for robotic navigation in construction contexts. The commands are entered into a Bayesian framework that will affect the robot's heuristic parameters. Results show a robust alignment of mission planning with different desired metric values. 
\par
 For future works, we plan to expand on this topic by introducing multiple prompts from different prompters. In this paper, we have theoretically proven the feasibility of this framework to account for multiple prompters. The goal is for the robot to understand the environment and the policy to implement in case of conflicting commands with appropriate confidence levels for each separate command, rather than to determine the task to execute in normal situations.

\section{Acknowledgments}
The presented work has been supported by the U.S. National Science Foundation (NSF) CAREER Award through grant No. CMMI 2047138, and grant No. DUE 1930546. The authors gratefully acknowledge the support from the NSF. Any opinions, findings, conclusions, and recommendations expressed in this paper are those of the authors and do not necessarily represent those of the NSF.

\bibliography{ISARC}

\end{document}